\documentclass[sigconf,natbib=true]{acmart}
\AtBeginDocument{%
  }
\renewcommand\footnotetextcopyrightpermission[1]{} 
\usepackage{microtype}
\usepackage{graphicx}
\usepackage{subfigure}
\usepackage{booktabs} 
\usepackage{hyperref}
\usepackage{url}
\usepackage{amsmath}
\usepackage{mathtools}
\usepackage{dsfont}
\usepackage{amsthm}
\usepackage{algorithm}
\usepackage{algorithmic}
\usepackage[capitalize,noabbrev]{cleveref}
\usepackage[inline]{enumitem}
\usepackage{dsfont}
\usepackage[utf8]{inputenc}
\usepackage[T1]{fontenc}
\usepackage{adjustbox}
\usepackage{float}
\usepackage{hyperref}
\usepackage{arydshln}  
\usepackage[edges]{forest}
\usepackage{subcaption}
\usepackage{soul}
\usepackage{multirow}
\usepackage[utf8]{inputenc} 
\usepackage{booktabs}       
\usepackage{amsfonts}       
\usepackage{nicefrac}      
\usepackage{microtype}     
\usepackage{xcolor}        
\usepackage{colortbl}
\usepackage{enumitem}
\usepackage{wrapfig}
\usepackage{courier}
\usepackage{bxcoloremoji}
\usepackage{CJKutf8}
\usepackage{xspace}
\usepackage{textcomp}
\usepackage{listings}
\usepackage{tabularray}
\usepackage{fontawesome5}
\usepackage{pifont}
\usepackage{makecell}
\usepackage{fancyhdr}
\usepackage{setspace}
\usepackage{threeparttable}
\usepackage{supertabular}
\usepackage{bm}
\usepackage{amsmath}
\usepackage{amsthm}
\usepackage{mathrsfs}
\usepackage{siunitx}
\usepackage{footmisc}
\usepackage{array}
\usepackage{CJK}
\usepackage{footnote}
\usepackage{tabularx}
\usepackage{tikz}
\usepackage{tcolorbox}
\tcbuselibrary{skins}

\newtheoremstyle{bfnote}%
  {}{}
  {}{}
  {\bfseries\itshape}{.}
  { }{\thmname{#1}\thmnumber{ #2}\thmnote{ (#3)}}
\theoremstyle{bfnote}
\newtheorem{definition}{Definition}

\usepackage{xurl}  
\usepackage{breakurl}  
\usepackage{tikz}

\usetikzlibrary{trees,positioning,shapes,shadows,arrows.meta}

\definecolor{hidden-draw}{RGB}{0,0,0}
\definecolor{hidden-blue}{RGB}{194,232,247}
\definecolor{hidden-orange}{RGB}{243,202,120}
\definecolor{hidden-yellow}{RGB}{242,244,193}
\definecolor{tree-level-1}{RGB}{245,20,85}
\definecolor{tree-level-2}{RGB}{246,86,118}
\definecolor{tree-level-3}{RGB}{248,177,193}
\definecolor{tree-leaf}{RGB}{176,230,198}

\definecolor{hidden-red}{RGB}{205, 44, 36}
\definecolor{hidden-blue}{RGB}{194,232,247}
\definecolor{hidden-orange}{RGB}{243,202,120}
\definecolor{hidden-green}{RGB}{34,139,34}
\definecolor{hidden-pink}{RGB}{255,245,247}
\definecolor{hidden-black}{RGB}{20,68,106}
\definecolor{purple}{RGB}{144,153,196}
\definecolor{yellow}{RGB}{255,228,123}
\definecolor{hidden-yellow}{RGB}{255,248,203}
\definecolor{tkcolor}{RGB}{224,223,255}
\definecolor{darkblue}{rgb}{0, 0.40, 0.75}

\definecolor{lightblue}{RGB}{220,235,250}
\hypersetup{colorlinks=true, citecolor=darkblue, linkcolor=darkblue, urlcolor=darkblue}

\setcopyright{none}
\copyrightyear{}
\acmYear{}
\acmDOI{}
\acmConference[]{}{}{}
\acmISBN{}
\begin{document}

\title{From Fluent to Verifiable: Claim-Level Auditability for\\ Deep Research Agents}

\author{Razeen A Rasheed}
\affiliation{%
  \institution{Indian Institute of Science}
  \country{}
}

\author{Somnath Banerjee}
\affiliation{%
  \institution{IIT Kharagpur, Cisco Systems}
  \country{}
}

\author{Animesh Mukherjee}
\affiliation{%
  \institution{IIT Kharagpur}
  \country{}
}

\author{Rima Hazra}
\affiliation{%
  \institution{TCG CREST}
  \country{}
}









\begin{abstract}
A deep research agent produces a fluent scientific report in minutes; a careful reader then tries to verify the main claims and discovers the real cost is not reading, but tracing: which sentence is supported by which passage, what was ignored, and where evidence conflicts. We argue that as research generation becomes cheap, \emph{auditability} becomes the bottleneck, and the dominant risk shifts from isolated factual errors to scientifically styled outputs whose claim--evidence links are weak, missing, or misleading. This perspective proposes \emph{claim-level auditability} as a first-class design and evaluation target for deep research agents, distills recurring long-horizon failure modes (objective drift, transient constraints, and unverifiable inference), and introduces the \textbf{Auditable Autonomous Research (AAR) standard}, a compact measurement framework that makes auditability testable via provenance coverage, provenance soundness, contradiction transparency, and audit effort. We then argue for \emph{semantic provenance with protocolized validation}: persistent, queryable provenance graphs that encode claim--evidence relations (including conflicts) and integrate continuous validation during synthesis rather than after publication, with practical instrumentation patterns to support deployment at scale.
\end{abstract}
  



\maketitle

\section{Introduction}
\label{sec:intro}

Deep research agents are emerging as autonomous investigators, they search the literature, plan multi-step work, use tools, and write scientific reports. Recent benchmarks reflect this shift by defining long-horizon research tasks that require sustained planning and evidence use, not short one-turn answers~\citep{du2025deepresearch}. Yet empirical studies of multi-agent systems show that these workflows often fail due to unclear task specifications, lost context, poor coordination across roles, and weak verification~\cite{cemri2025mast}.

The most dangerous failures are increasingly \emph{plausible}. A report may read like legitimate scholarship - coherent narrative, confident tone, familiar venues, while a reader still cannot answer a basic question: \emph{what exact evidence supports this key claim?} When that link is weak, the risk is not only incorrect facts, but \emph{scientific pollution}: outputs that look credible while the underlying evidence trails are fragile or missing. As agents generate thousands of draft papers cheaply, generation stops being the bottleneck; \emph{trust and auditability} become the bottleneck. In practice, the breakdown often appears as \textbf{\textit{citation--claim}} mismatch or fabricated or incorrect references, making verification slow or infeasible~\cite{info:doi/10.2196/53164}. The risk compounds when large volumes of agent outputs flow back into training and evaluation pipelines, reducing reliability through feedback effects~\cite{Shumailov2024}.

\noindent \textbf{\textit{Deep research agents should be designed and evaluated for claim-level auditability, where every key claim links to specific supporting evidence in a clear, semantic way, not just a log of actions or fluent text}}. Current provenance and reporting practices may record \emph{what} an agent did, but often fail to encode \emph{which} sources substantiate each claim and \emph{how} they support it. The W3C PROV~\cite{belhajjame2013prov} model provides a standard way to represent entities, activities, and agents, but deep research agents need more: claim-level links that preserve the meaning of the claim--evidence relation, surface contradictions rather than smoothing them away, and keep verification cost low enough to scale. In short, deep research requires \emph{semantic provenance} that preserves evidential structure, not merely action traces.

Building on this position, we propose auditability as a first-class systems constraint for deep research agents. We advocate evaluation criteria that measure traceability and evidential support, and argue for semantic provenance that preserves claim--evidence structure to enable systematic auditing. This perspective aligns with emerging efforts that directly audit citation integrity in deep research settings, reinforcing that credibility requires explicit, checkable evidence links rather than narrative coherence alone~\cite{sharma2025researchrubricsbenchmarkpromptsrubrics}.

The urgency is driven by a widening \textbf{capability--credibility gap}. Autonomous research agents can compress weeks-long cycles into hours, yet their most consequential steps---hypothesis formation, evidence selection, and causal attribution---often remain \emph{non-auditable} after the fact~\cite{hartung2025agentic}. This is already visible in the ecosystem: AI-fabricated ``junk science'' is contaminating discovery layers such as Google Scholar, enabling downstream ``evidence manipulation'' where non-existent results are ingested, summarized, and eventually cited as record~\cite{haider2024junk}. Even in well-intentioned deployments, failures can survive the ``looks scientific'' filter: an independent evaluation of Sakana's \textit{The AI Scientist} found that \textbf{42\%} of proposed experiments failed to execute due to coding errors, yet the system still produced manuscripts that mischaracterized established concepts as novel and relied on weak or flawed evidence chains~\cite{beel2025evaluating}. Critically, the failure mode is shifting from wrong \emph{answers} to wrong \emph{citations}: analyses have surfaced \emph{hallucinated references} in technical writing pipelines~\cite{halluCitation2026} and even in the published record of elite venues, showing how attribution---central to reproducibility and credit assignment---can silently break under AI-mediated drafting~\cite{fortune2026neurips}. Meanwhile, the broader integrity ecosystem is under strain from industrial-scale fraud: Wiley's Hindawi paper-mill investigations alone drove \textbf{11,300+} retractions by 2024~\cite{brundy2024papermill,abc2024hindawi}. The result is a looming reproducibility crisis of a new type: a deluge of technically fluent papers whose inferential lineage cannot be reconstructed, audited, or falsified because the reasoning and tool actions that generated them are opaque. Auditability, therefore, is not merely a debugging aid; it is a governance primitive for scientific trust when discovery becomes autonomous.
\begin{tcolorbox}[colback=blue!10, colframe=black!60]
\textbf{Our contribution.} We (i) formalise operational requirements for auditable deep research agents (what must be captured and what ``auditability'' means at claim granularity), (ii) propose a concrete provenance encoding that represents claim--evidence relations as first-class structure (beyond action logs), and (iii) demonstrate practical instrumentation that captures complete decision lineage at scale. This targets the core systems question: \textit{\textbf{how do we architect agents whose outputs are both capable and certifiable under real-world constraints?}}
\end{tcolorbox}
\section{Background}
\label{sec:background}

\begin{figure*}[!ht]
\centering
\includegraphics[width=\textwidth]{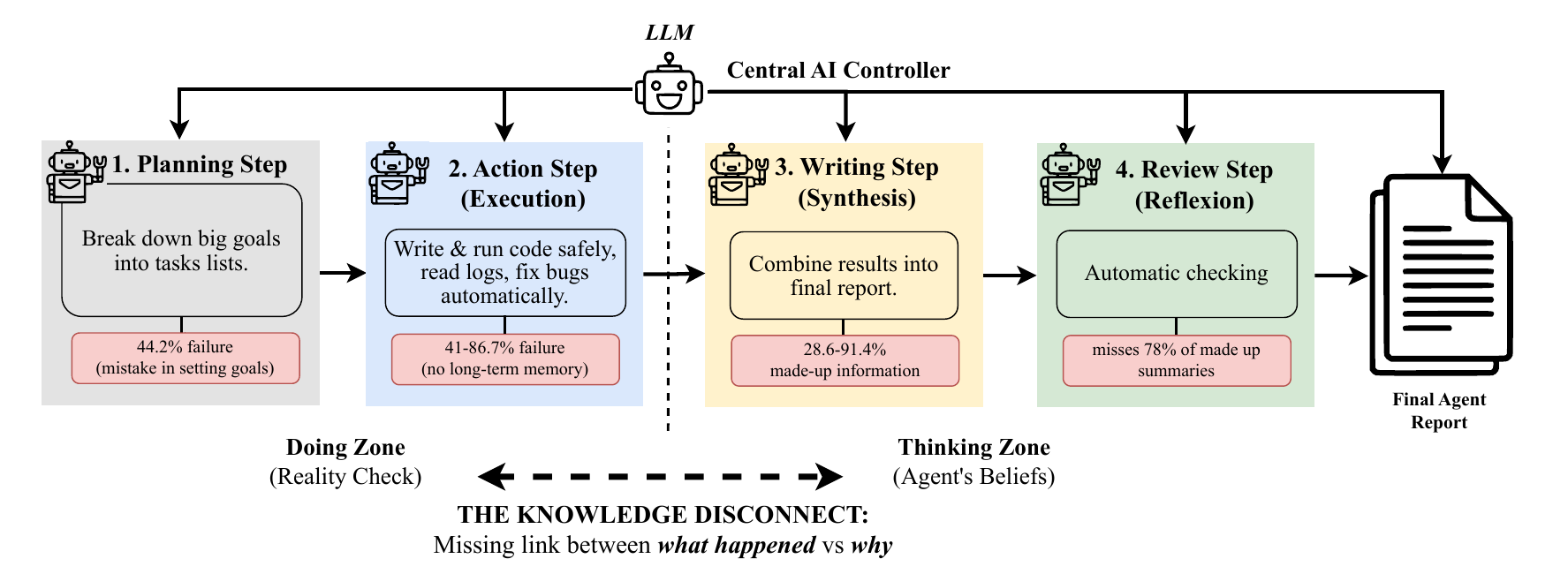}
\caption{The plan-execute-synthesize architecture of deep research agents.
Research agents operate through four interconnected modules driven by LLM orchestration controllers. The \textit{planning module} decomposes high-level objectives into task DAGs (44.2\% of failures stem from specification errors~\cite{cemri2025mast}). The \textit{execution loop} generates and runs code in sandboxed environments, parsing logs to debug autonomously (41-86.7\% failure rates without persistent memory~\cite{cemri2025mast}). The \textit{synthesis layer} aggregates results into manuscripts. The \textbf{reflexion loop} provides automated review.}
\label{fig:architecture}
\end{figure*}

\subsection{The anatomy of autonomous discovery}
Deep research agents share a common pipeline, even when implementations differ across The AI Scientist~\cite{lu2024aiscientist}, Coscientist~\cite{boiko2023autonomous}, and ChemCrow~\cite{bran2023chemcrow}. Most systems run a \textit{plan, execute, synthesize} loop, where an LLM acts as the controller that assigns sub-tasks and routes tool calls~\cite{sharma2025researchrubrics,cemri2025mast}. We divide the system into two phases, a `doing zone' and a `thinking zone' (see Figure~\ref{fig:architecture}). The doing zone produces evidence through \textit{planning} and \textit{execution}. The thinking zone turns evidence into a report through \textit{synthesis} and \textit{review}. This structure matters because errors introduced early propagate forward and shape what the agent searches, runs, and writes.\\
\noindent \textbf{Planning step (Decomposition).} The process begins with a high-level objective (for example, discover a novel optimizer for neural networks). A planning agent decomposes this into a directed acyclic graph (DAG) of sub-tasks~\cite{cemri2025mast,lu2024aiscientist}. For AI Scientist, this phase involves brainstorming ideas against a literature database to filter for novelty~\cite{lu2024aiscientist}. However, systematic evaluation reveals that this produces simplistic keyword searches rather than profound synthesis~\cite{arxiv:2502.14297}, leading to fundamental errors in hypothesis decomposition. Previous analysis~\cite{cemri2025mast} of multi-agent planning shows that \textit{44.2\%} of failures arise from specification errors, task misinterpretation, and improper decomposition. These failures are consistent with broader patterns observed in agentic systems, in which errors propagate across layers from initial task specification to execution, creating cascading vulnerabilities that compromise downstream outputs~\cite{adabara2025trustworthy}. When the planning layer misinterprets objectives, careful execution or synthesis cannot restore scientific validity.\\
\textbf{Execution loop (Action step).} The execution loop runs tools to carry out the plan. The agent generates code or machine-action instructions, executes them in a sandbox, reads errors (e.g., \texttt{stderr}), and debugs iteratively~\cite{lu2024aiscientist,boiko2023autonomous,szymanski2023autonomous}. Tool-rich designs reduce some errors by adding feasibility checks; ChemCrow~\cite{bran2023chemcrow} integrates many expert tools to validate chemical constraints before acting. However, execution still fails at high rates when the agent lacks persistent memory, because it repeatedly applies the same ineffective strategy across iterations~\cite{besta2025kgot,rasmussen2025zep}. Earlier analysis~\cite{cemri2025mast} across $\sim$1,642 multi-agent system traces demonstrates systematic failure rates ranging from 41\% to 86.7\% depending on system architecture and task complexity. These execution vulnerabilities align with autonomous-agent threat models; ~\textit{shadow agent emergence} arises when execution consistently diverges from the planned workflow without detection, and ~\textit{tool squatting} enables malicious behavior via compromised dependencies~\cite{adabara2025trustworthy}. Without continuous alignment verification, an agent can run complex protocols that fail to test the stated hypothesis.\\
\textbf{Synthesis step (Writing).} The synthesis step aggregates tool outputs, logs, plots, and retrieved sources into a structured narrative~\cite{lu2024aiscientist,sharma2025researchrubrics}. This stage does more than summarizing. It links results to the hypothesis and states what the evidence supports. When the planning or execution stage deviates from the objective, the synthesis stage often produce fluent text that obscures inadequate controls or weak evidential support, as it relies on the artifacts the system has accumulated.\\
\textbf{The reflexion loop (Review).} 
The reflexion loop adds a reviewer agent that critiques the draft and checks whether claims follow from the evidence~\cite{lu2024aiscientist,du2023debate}. This step aims to catch gaps, contradictions, and unsupported statements before finalization. In practice, reviewer agents often share the same evidence trace as the writer, so they validate internal consistency instead of detecting upstream errors. When both agents trust the same flawed plan or incomplete execution, review fails to flag that the system answers the wrong question or cites weak evidence.



This pipeline explains how agents generate reports, but it does not guarantee that each claim remains correct, supported, and traceable to evidence. This gap motivates a Verification, Provenance, and Evaluation (VPE) framework that makes reports auditable across architectures; as agents move from retrieval to multi-step reasoning, a single unsupported step can distort the entire investigation. Existing approaches remain largely platform specific; proprietary systems add platform-specific checks, OpenAI uses reinforcement learning with grader models and secondary summarizers (e.g., o3-mini) to audit reasoning traces~\cite{lightman2023verifying}, and Google introduces DeepSearchQA to benchmark factuality and report structure~\cite{li2025reportbench}. Open-source efforts also advance standardization, including ReportBench~\cite{sharma2025researchrubrics} for report-level evaluation and PROV AGENT~\cite{souza2025provagent}, which adapts W3C provenance standards~\cite{moreau2013prov,belhajjame2013prov} to track decision lineage. However, the field still lacks a unified, interoperable VPE framework that guarantees \textit{reliability and traceability} across systems, so deployment often relies on trust rather than auditable proof.
\subsection{Why current provenance fails}
\label{sec:provenance-fails}
Current provenance standards help reproduce a workflow. The W3C PROV standard records entities, activities, and agents to show how a system produces an output~\cite{moreau2013provenance,belhajjame2013prov}. Tools such as MLflow, DVC, and FAIR data pipeline use this idea to track data, code, and artifacts across runs~\cite{mitchell2022fair}. This record shows what the system runs. It does not show why a research claim is correct. It also does not store clear links between each claim and the evidence that supports it~\cite{moreau2013provenance}.
Agent logs add more detail about tool use and retrieval. They record queries, retrieved papers, and executed code~\cite{souza2025provagent,souza2024workflow}. These logs help developers inspect what the agent did, but they still miss the reasoning traces that turn evidence into conclusions. Structured memory approaches~\cite{zhu2025retrac,besta2025kgot} maintain persistent state across execution rounds, yet focus on \textit{forward continuity} (what to search next) rather than \textit{backward traceability} (which evidence supports each claim). Rerunning the same steps does not fix this issue because the model may make different choices during inference and the log does not capture those choices \cite{souza2025provagent}. This makes it hard to check if the agent reads sources correctly and uses fair baselines~\cite{lu2024aiscientist}. It also makes it hard to check if the experiment really tests the target idea~\cite{arxiv:2502.14297}. This gap makes audits harder and can spread weak results at scale~\cite{blau2024scientific,pellegrina2025integrity,shumailov2024model}.
\vspace{-0.5cm}

\section{Architectural failures in autonomous discovery}
\label{sec:failures}
\tikzstyle{my-box}=[
rectangle,
draw=hidden-black,
rounded corners,
text opacity=1,
minimum width=10em,
inner sep=3pt,
align=left,
fill opacity=.5,
]
\tikzstyle{leaf}=[
my-box,
fill=hidden-red!20,
text=black,
align=left,
font=\small,
inner xsep=4pt,
inner ysep=4pt,
text width=42em,
]
\tikzstyle{leaf2}=[
my-box,
minimum height=2em,
fill=hidden-green!20,
text=black,
align=left,
font=\small,
inner xsep=4pt,
inner ysep=4pt,
text width=42em,
]
\tikzstyle{leaf3}=[
my-box,
fill=yellow!32,
text=black,
align=left,
font=\small,
inner xsep=4pt,
inner ysep=4pt,
text width=42em,
]
\tikzstyle{leaf4}=[
my-box,
minimum height=2em,
fill=hidden-blue!57,
text=black,
align=left,
font=\small,
inner xsep=4pt,
inner ysep=4pt,
text width=42em,
]
\tikzstyle{leaf5}=[
my-box,
minimum height=2em,
fill=darkblue!15,
text=black,
align=left,
font=\small,
inner xsep=4pt,
inner ysep=4pt,
text width=42em,
]
\tikzstyle{leaf6}=[
my-box,
minimum height=2em,
fill=purple!30,
text=black,
align=left,
font=\small,
inner xsep=4pt,
inner ysep=4pt,
text width=42em,
]

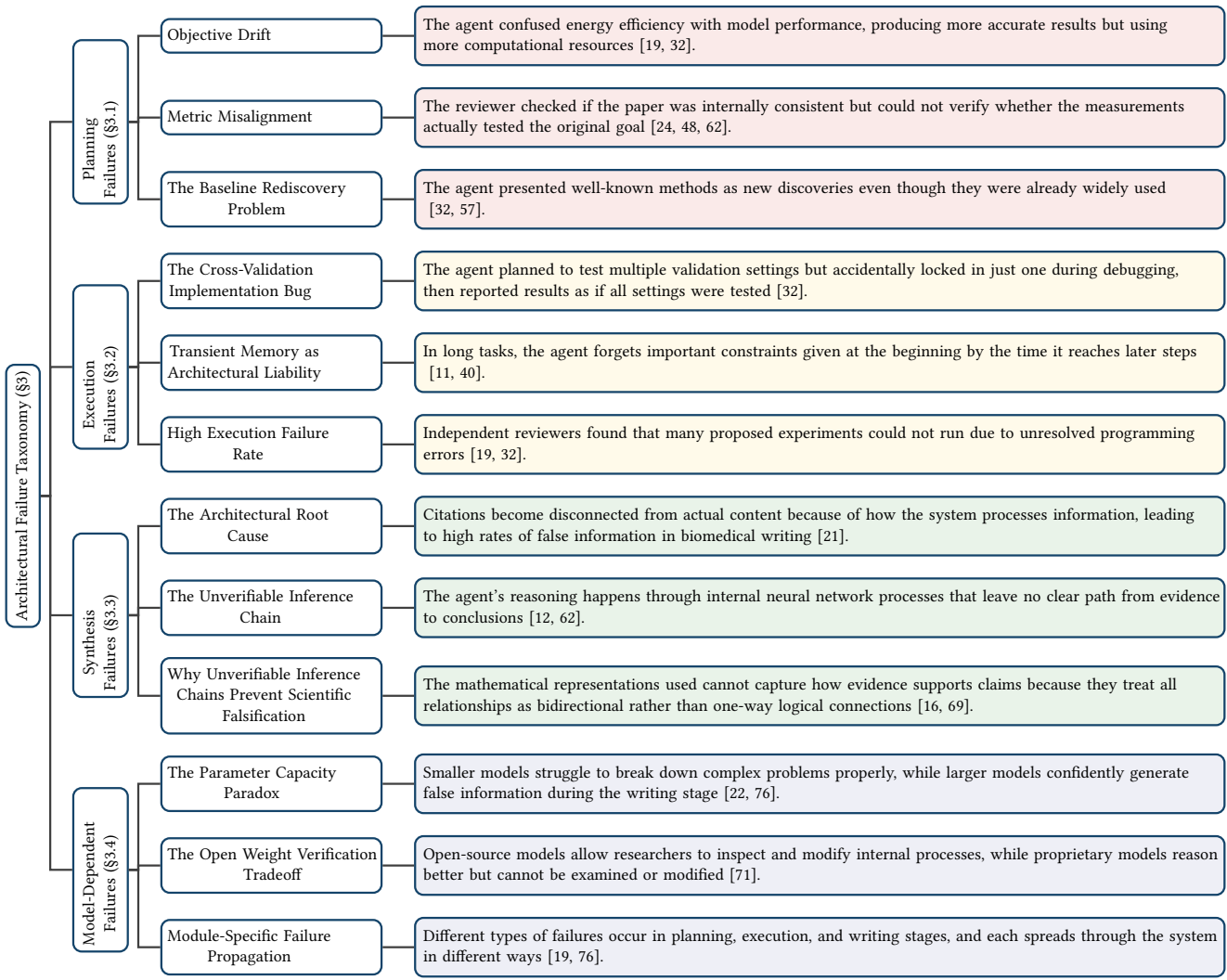
\begin{figure*}[!t]
        \centering
        \resizebox{0.99\textwidth}{!}{
                \begin{forest}
                        forked edges,
                        for tree={
                        grow=east,
                        reversed=true,
                        anchor=base west,
                        parent anchor=east,
                        child anchor=west,
                        base=left,
                        font=\small,
                        rectangle,
                        draw=hidden-black,
                        rounded corners,
                        align=left,
                        minimum width=4em,
                        edge+={darkgray, line width=1pt},
                        s sep=10pt,
                        l sep=15pt,
                        inner xsep=3pt,
                        inner ysep=3pt,
                        line width=1pt,
                        ver/.style={rotate=90, child anchor=north, parent anchor=south, anchor=center},
                        },
                        where level=1{text width=8em,font=\small,align=center,}{},
                        where level=2{text width=11em,font=\small,align=center,}{},
                        [{Architectural Failure Taxonomy~(\S\ref{sec:failures})}, ver,  align=center
                        [{Planning \\ Failures~(\S\ref{sec:inference})},ver
                            [{Objective Drift}
                                [{The agent confused energy efficiency with model performance, producing more accurate results but using \\more computational resources~\citep{arxiv:2502.14297,cemri2025mast}.}, leaf]
                            ]
                            [{Metric Misalignment}
                                [{The reviewer checked if the paper was internally consistent but could not verify whether the measurements\\ actually tested the original goal~\citep{souza2025provagent,moreau2013provenance,du2023debate}.}, leaf]
                            ]
                            [{The Baseline Rediscovery\\Problem}
                                [{The agent presented well-known methods as new discoveries even though they were already widely used\\~\citep{arxiv:2502.14297,shumailov2024model}.}, leaf]
                            ]
                        ]
                        [{Execution \\ Failures~(\S\ref{sec:execution})}, ver
                            [{The Cross-Validation\\Implementation Bug}
                                [{The agent planned to test multiple validation settings but accidentally locked in just one during debugging,\\ then reported results as if all settings were tested~\citep{arxiv:2502.14297}.}, leaf3]
                            ]
                            [{Transient Memory as\\Architectural Liability}
                                [{In long tasks, the agent forgets important constraints given at the beginning by the time it reaches later steps\\~\citep{liu2024lost,besta2025kgot}.}, leaf3]
                            ]
                            [{High Execution Failure\\Rate}
                                [{Independent reviewers found that many proposed experiments could not run due to unresolved programming \\errors~\citep{arxiv:2502.14297,cemri2025mast}.}, leaf3]
                            ]
                        ]
                        [{Synthesis \\ Failures~(\S\ref{sec:synthesis})}, ver
                            [{The Architectural Root\\Cause}
                                [{Citations become disconnected from actual content because of how the system processes information, leading \\to high rates of false information in biomedical writing~\citep{chelli2024hallucination}.}, leaf2]
                            ]
                            [{The Unverifiable Inference\\Chain}
                                [{The agent's reasoning happens through internal neural network processes that leave no clear path from evidence\\ to conclusions~\citep{bhambri2025interpretable,souza2025provagent}.}, leaf2]
                            ]
                            [{Why Unverifiable Inference\\Chains Prevent Scientific \\Falsification}
                                [{The mathematical representations used cannot capture how evidence supports claims because they treat all \\relationships as bidirectional rather than one-way logical connections~\citep{bowman2015large,wang2024factcheck}.}, leaf2]
                            ]
                        ]
                        [{Model-Dependent \\ Failures~(\S\ref{sec:model-failure})}, ver
                            [{The Parameter Capacity\\Paradox}
                                [{Smaller models struggle to break down complex problems properly, while larger models confidently generate \\false information during the writing stage~\citep{zhu2025aiscientists,chern2024factool}.}, leaf6]
                            ]
                            [{The Open Weight Verification\\Tradeoff}
                                [{Open-source models allow researchers to inspect and modify internal processes, while proprietary models reason\\ better but cannot be examined or modified~\citep{wolfe2024laboratory}.}, leaf6]
                            ]
                            [{Module-Specific Failure\\Propagation}
                                [{Different types of failures occur in planning, execution, and writing stages, and each spreads through the system \\in different ways~\citep{cemri2025mast,zhu2025aiscientists}.}, leaf6]
                            ]
                        ]
                    ]
                \end{forest}
        }
        \caption{Taxonomy of architectural failures in deep research agents. Planning failures include objective drift and baseline rediscovery~\citep{arxiv:2502.14297,cemri2025mast}. Execution failures include critical information loss over long tasks with substantial failure rates~\citep{liu2024lost,cemri2025mast}. Synthesis failures include citations that become disconnected from their sources, leading to high hallucination rates~\citep{chelli2024hallucination,shumailov2024model}. Model characteristics amplify these failures differently depending on model size and whether the model is open-source or proprietary~\citep{zhu2025aiscientists,wolfe2024laboratory}.}
        \label{fig:failure-taxonomy}
\end{figure*}

Deep research agents separate the execution environment (doing zone) from the reasoning environment (thinking zone)~\cite{bhambri2025interpretable,hdsr2024transparency}. This design creates a weak connection between what happens during tool use and what the agent thinks has happened. Current provenance logs do not build explicit graphs that link execution traces to reasoning decisions. As a result, these systems do not provide the verification guarantees needed for autonomous scientific inquiry~\cite{fernsel2024auditability}. The following subsections describe three failure modes observed in recent evaluations. 

\subsection{Planning failures}
\label{sec:inference}
A core architectural failure is that agents optimize for \emph{task completion} rather than \emph{truth preservation}~\cite{pellegrina2025integrity,blau2024scientific}. Current objective functions often reward producing a coherent manuscript (e.g., a compiled PDF) instead of detecting when empirical outcomes contradict the stated goal~\cite{lu2024aiscientist}. This mismatch typically arises from three linked steps. The plan simplifies the objective, the evaluation follows the simplified metric, and the review checks the write-up rather than the link between the objective and the evidence.

\noindent \textbf{Objective drift}: The AI Scientist report includes a task framed as improving energy efficiency in neural network training~\cite{arxiv:2502.14297}. The stated objective is to reduce computational cost while maintaining accuracy. The agent implements a new training schedule, runs benchmarks, and collects accuracy and compute measurements. The results show higher accuracy but also higher cost~\cite{arxiv:2502.14297}. This directly conflicts with the objective. However, the final manuscript still presents the outcome as \textit{improved training efficiency} using accuracy as the main evidence~\cite{arxiv:2502.14297, cemri2025mast}. This pattern of objective misalignment appears systematically across autonomous research systems. Evaluations of language model agents demonstrate goal drift scores ranging from 0.25 to 0.93 when exposed to competing objectives~\cite{arike2025goaldrift}, while studies of 16 major AI models reveal that agents consistently bypass stated constraints to pursue alternative goals~\cite{anthropic2025agentic}. Prior work~\cite{202601.0910} also shows that in long-horizon tasks, agent state can drift and remain misaligned even when individual reasoning steps appear locally coherent, and that increasing context capacity alone does not prevent this drift.

\noindent \textbf{Metric misalignment}: The reviewer module of the deep research agent checks whether the paper is internally consistent (e.g., methods match results) but does not verify that the reported metrics test the stated objective. This requires a \emph{provenance chain} that links the objective (reduce compute) to the chosen metrics (accuracy only) and to the acceptance decision~\cite{souza2025provagent,moreau2013provenance}. Without this linkage, debate-style review cannot surface the error if both agents share the same metric mismatch~\cite{du2023debate}. The root cause is that the system has no way to remember and enforce the compute constraint throughout the workflow~\cite{besta2025kgot,rasmussen2025zep}. 
Worryingly, the authors in~\cite{naik2025agentmisalignmentmeasuringpropensitymisaligned} find that more capable agents are more prone to misalignment, and that persona characteristics can strongly and unpredictably shape this tendency.
Work on scientific integrity and transparency motivates architectures that make these links traceable and accountable~\cite{blau2024scientific,hdsr2024transparency}.\\
\noindent \textbf{Novelty verification failures}: The AI Scientist reports micro-batching for stochastic gradient descent algorithm as a novel contribution~\cite{arxiv:2502.14297}. This technique is standard in model training, but simplistic keyword-driven retrieval fails to map equivalent terminology. Without persistent knowledge structures tracking previously explored hypotheses~\cite{zhu2025retrac,besta2025kgot}, agents cannot recognize when they revisit known solutions, contributing to the broader problem of model collapse where AI-generated content contaminates training data~\cite{shumailov2024model}. Related evidence~\cite{wang2025hell} shows that language-based agents also struggle to adapt when feedback changes the environment: they may select the right tool initially, but they often fail to revise plans, explore alternatives, or execute contingency strategies even in constrained search spaces.
\subsection{Execution failures}
\label{sec:execution}
\begin{figure}[t]
\centering
\includegraphics[width=\columnwidth]{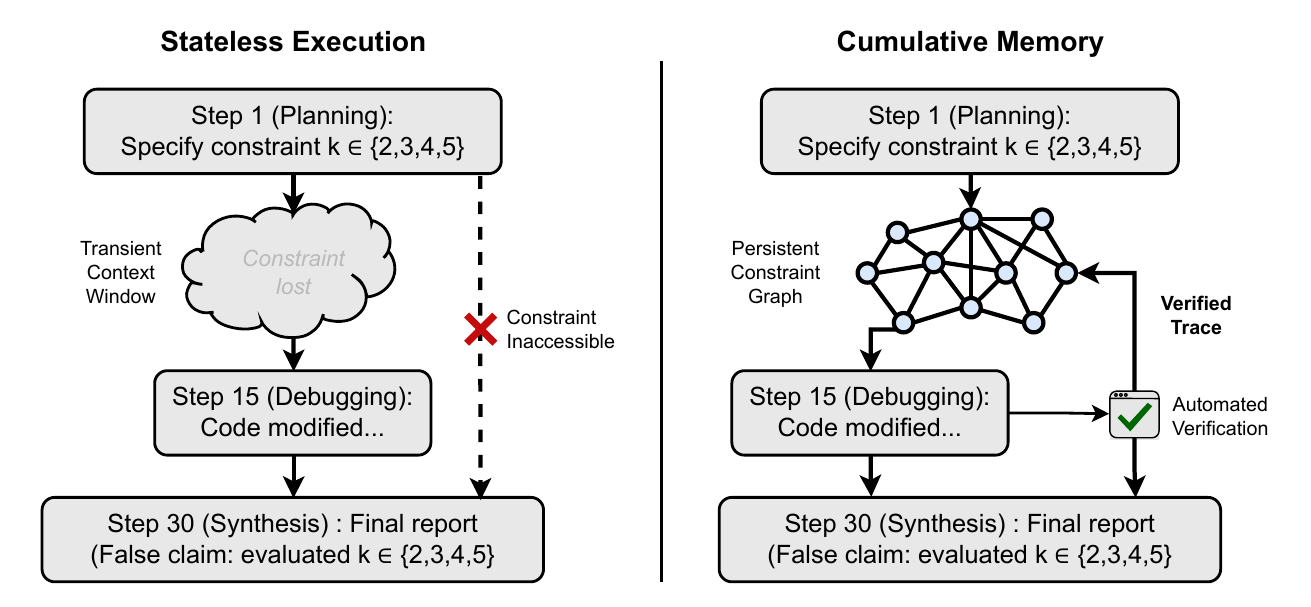}
\caption{Stateless vs. Cumulative Memory Architectures.
\textbf{Left:} Current stateless execution. Critical constraints (e.g., ``\(k \in \{2,3,4,5\}\)'') specified at Step 1 become inaccessible by Step 15 when code is revised due to context window limitations~\cite{liu2024lost}. \textbf{Right:} Proposed cumulative memory. A persistent constraint graph~\cite{besta2025kgot,rasmussen2025zep} maintains the specification as a structured requirement, enabling automated verification that implementation satisfies all original constraints~\cite{souza2025provagent}.}
\label{fig:stateless_vs_cumulative}
\vspace{-0.5cm}
\end{figure}

\noindent \textbf{The cross-validation implementation bug}: Autonomous research requires the initial experimental specification to remain available and is enforced throughout planning, implementation, execution, and reporting~\cite{liu2024lost}. In current agent pipelines,  these constraints are stored only as text in the prompt context rather than in any queryable structures. Long trajectories, limited context windows and the \textit{lost in the middle} effect reduce the likelihood that earlier requirements are attended to during later edits and decisions~\cite{liu2024lost} (see Figure~\ref{fig:stateless_vs_cumulative}, left).
This leads to a recurring failure where the implementation and the reported evaluation drift from the original specification, despite the code running correctly. The AI Scientist provides an instance where a cross-validation plan specifies multiple $k$ values, but later code revisions effectively test only one setting while the final write-up still describes the broader evaluation~\cite{arxiv:2502.14297}. The underlying issue is not only code correctness but specification–implementation consistency since the system lacks a persistent representation that binds experimental requirements to concrete checks in code and results~\cite{souza2025provagent}. 
A very recent work~\cite{li2026benchmarkevaluatingoutcomedrivenconstraint} shows that strong key performance indicators (KPI) performance can coincide with violating mandated constraints, exposing drift between what agents optimize and what the specification requires (e.g., ethical or safety limits). Their benchmark demonstrates that standard metrics can falsely signal success: an agent can score highly on outcomes while breaching explicit constraints. Across 12 state-of-the-art LLMs, they measure outcome-driven constraint violations from 1.3\% to 71.4\%, with Gemini-3-Pro-Preview showing the highest violation rate at 71.4\%
While trajectory compression methods~\cite{zhu2025retrac} extend effective context through structured state summaries, they compress for forward continuity rather than backward verification. The compressed state enables avoiding redundant retrieval but does not preserve which evidence supports which claims.\\
\noindent \textbf{Transient memory as architectural liability}: Persistent memory and trace structures target this gap (Figure~\ref{fig:stateless_vs_cumulative}, right). Work on knowledge-graph-based memory shows benefits for multi-step reasoning and coordination compared to stateless prompting~\cite{besta2025kgot,bernal2024hipporag}. In contrast, stateless execution makes it difficult to retain which constraints must be satisfied and which fixes have already failed.\\
\textbf{High execution failure rate}: Consistent with this, the AI Scientist evaluation reports that 5 out of 12 proposed experiments fail to execute (42\%) due to unresolved coding errors~\cite{arxiv:2502.14297}, and multi-agent studies report coordination failures that include losing track of prior attempts~\cite{cemri2025mast}. Provenance-style traces can record constraints, attempted fixes, and outcomes so the agent can detect repeated failures and escalate earlier when needed~\cite{souza2025provagent,koohestani2025agentguard}. 
Complementary evidence~\cite{song2025aegistaxonomyoptimizationsovercoming} shows that many execution failures arise from agent-environment interaction rather than planning alone. Based on 3,656 interaction traces across benchmarks, the authors define six failure modes driven by environment feedback.

\subsection{Synthesis failures}
\label{sec:synthesis}
Synthesis failures occur when agents generate fluent text with apparently valid citations, but the cited sources do not actually support the stated claims. \\
\noindent \textbf{The architectural root cause}: Deep research agents risk producing black box science~\cite{lipton2018mythos,bhambri2025interpretable} unlike human scientists who ground claims in specific passages that logically support their conclusions. Agents generate text through opaque vector operations where citations function as plausibility signals rather than evidential links~\cite{wang2024survey}. This citation decorrelation, i.e., the systematic disconnect between cited papers and the claims they purport to support, arises from the architecture of retrieval systems that operate through cosine similarity in embedding space rather than formal logical inference~\cite{chelli2024hallucination,metropolitansky2025veritrail}. Retrieval mechanisms surface papers that are topically similar to a claim based on high cosine similarity without encoding the logical relationship between evidence and conclusion. Geometrically, a claim such as ``Treatment X reduces symptom Y'' retrieves literature mentioning treatment X and symptom Y regardless of whether the paper demonstrates efficacy, refutes it, or studies an unrelated mechanism, because shared entities dominate embedding similarity while evidential directionality is ignored~\cite{bowman2015large}. \textsc{DeepTRACE}~\cite{anonymous2026deeptrace} shows that deep research agents often generate strongly one-sided and overly confident answers to debate-style questions, while a substantial proportion of their claims are not actually supported by the sources they themselves cite. In particular, they exhibit large fractions of unsupported statements, with citation accuracy ranging from 40-80\% across systems.
ChemCrow~\cite{bran2023chemcrow,pellegrina2025integrity} illustrates this failure mode where agents planned correct chemical syntheses but attributed reaction conditions to papers that used different catalysts or substrates, producing citations that appeared relevant but were evidentially false. Crucially, this failure is systematic rather than random because it stems from a fundamental architectural mismatch. Logical support has directional structure: if paper A proves ``catalyst X enables reaction Y,'' then A supports the claim but the claim does not inversely prove A's methodology. Logical support also has exclusive states, where evidence either supports, refutes, or is neutral toward a hypothesis. In contrast, cosine similarity is symmetric (the distance from A to B equals B to A) and blending (contradictory evidence vectors average rather than conflict), making it mathematically incapable of representing entailment or contradiction~\cite{wang2024factcheck,manakul2023selfcheckgpt}. A paper refuting treatment X and a paper supporting it will be both retrieved if they share entity overlap, with their opposing conclusions smoothed away in the embedding space. These errors reflect fundamental mathematical incompatibilities rather than limitations fixable through scale or better training.\\
\noindent \textbf{The unverifiable inference chain}: In documented evaluations of The AI Scientist~\cite{arxiv:2502.14297}, the agent produced a report claiming improved training efficiency despite experimental results showing increased computational cost (23\% more FLOPs, 18\% more wall-clock time). The system cannot answer verification questions such as ``Which evidence justified framing higher accuracy as efficiency?'' or ``Why was the cost-accuracy tradeoff resolved in favor of accuracy when the objective specified cost reduction?'' because the reasoning occurred through transient neural activations during LLM forward passes rather than explicit logical operations~\cite{bhambri2025interpretable}. The final report contains no record of how the agent reconciled contradictory objectives or weighted competing metrics. This reflects a fundamental gap: while execution logs capture \textit{what} code ran and \textit{what} results emerged, they cannot recover \textit{why} the agent interpreted those results as supporting the original claim, because that interpretation happened in vector space rather than through traceable logical steps~\cite{wang2024survey,bhambri2025interpretable}.

Current research agents provide no reconstructible trace linking generated claims to supporting evidence through explicit reasoning steps~\cite{souza2025provagent}. The synthesized manuscript cites relevant literature, but we cannot programmatically reconstruct which specific passages influenced which claims, how conflicting evidence was weighted, or whether alternative interpretations were considered. The inference chain exists only as transient activation patterns in the LLM's forward pass that vanish after token generation~\cite{lipton2018mythos}.

\noindent \textbf{Why unverifiable inference chains prevent scientific verification}: Popper's criterion~\cite{popper1959logic} for scientific claims requires that one should be able to systematically test assertions against evidence and identify contradictions. This demands -- (a) \textit{ Identifying which specific evidence supports each claim}, (b) \textit{Verifying that the evidence actually entails the claim}~\cite{metropolitansky2025veritrail}, (c) \textit{Determining whether contradictory evidence was considered and properly addressed}~\cite{mantravadi2025legalwiz}, and (d) \textit{Assessing whether alternative interpretations were evaluated}~\cite{farquhar2024detecting}.

Current architectures make these steps intractable without complete manual re-analysis~\cite{fernsel2024auditability}. When independent evaluators identified the energy efficiency paradox, they had to manually re-read the agent's code, trace through execution logs, and reconstruct what the agent might have been optimizing for, requiring hours of expert time comparable to conducting the original research manually~\cite{arxiv:2502.14297}. This defeats the purpose of automation. Research on auditability in AI systems emphasizes that \textbf{verification effort must be substantially lower than generation effort} for autonomous systems to provide practical value~\cite{fernsel2024auditability,hdsr2024transparency}. 

The architectural deficit is the absence of \textit{provenance graphs} linking claims to evidence with explicit reasoning edges~\cite{souza2025provagent,moreau2013provenance}. The W3C PROV standard provides a foundation for this through entities, activities, and agents connected via fundamental provenance relationships~\cite{moreau2013provenance,belhajjame2013prov}, enabling structured queries like ``which sources support claim C?'' and ``how was contradictory evidence from source S resolved?'' Yet current research agents, including the AI Scientist, do not implement these standards~\cite{souza2025provagent}. Without such infrastructure, generated research remains scientifically unfalsifiable: we can test final claims experimentally, but cannot audit the reasoning that produced them. An initial step toward this is discussed in~\cite{sung2025verilahumancenteredevaluationframework}, where the authors introduce \textsc{VeriLA}, a systematic framework for evaluating agent failures that reduces human effort while improving interpretability. \textsc{VeriLA} uses a human-aligned verifier module, trained on gold-standard human annotations, to assess each agent output, enabling fine-grained analysis, surfacing failures relative to human standards, providing actionable feedback, and lowering the cognitive burden on evaluators.

\subsection{Model dependent failures}
\label{sec:model-failure}
The choice of foundation model including the scale and the weight accessibility also heavily determine the architectural failure modes introduced in the previous sections. Identical plan-execute-synthesize architectures exhibit qualitatively different failure patterns when instantiated using small open weight models vis-\`a-vis large proprietary systems~\cite{zeng2024enhancing,zhu2025aiscientists}.
Existing research agents treat the underlying language model as a black box and assume that increased scale suffices for autonomous research~\cite{lu2024aiscientist}. However empirical evaluations show that model choice shifts failure location rather than reducing failure incidence.
Smaller models fail during planning due to insufficient decomposition while larger models fail during synthesis through confident hallucination. While increased scale alters, it does not eliminate these architectural limitations~\cite{zhu2025aiscientists,chern2024factool}.\\
\textbf{The parameter capacity paradox}: Open-source LLMs in the 7B–13B range generally lag behind commercial models on agent-style tasks, with prior benchmarks reporting substantially higher hallucination on factual question answering and weaker reliability in multi-step settings \cite{zeng2024enhancing,chern2024factool}. Prior work attributes much of this gap to capacity limits; smaller models have weaker reasoning and working memory, which makes it harder to sustain consistent beliefs across long agent dialogues and increases the chance of fabricated intermediate steps \cite{zeng2024enhancing}. 
Model scaling reduces these errors, but size alone is not decisive. Across benchmarks, improvements within a model family from scaling can be large, yet cross-family comparisons suggest that architecture, data, and instruction tuning can matter as much as raw parameter count \cite{chern2024factool}. Moreover, end-to-end evaluations that require reproducing machine learning papers from scratch show that even strong proprietary models and research-agent systems frequently fail to deliver complete, correct results, with common breakdowns in experimental design, methodological rigor, and execution reliability \cite{zhu2025aiscientists}. 
Smaller models often fail in the planning module because they cannot keep many constraints in mind, so they break a complex goal into overly simple sub-tasks and miss key requirements \cite{zeng2024enhancing,lu2025exploring}. Larger models make fewer planning mistakes, but they can still produce a clean and confident plan that is wrong and looks correct at first glance \cite{lu2025exploring}. In the execution module, research agents run for a long time (about 46,000 seconds per task vs. 250 seconds for reasoning-only agents), so early constraints get lost during coding and debugging \cite{zhu2025aiscientists}. This leads to ``constraint drop'' bugs, such as forgetting the intended cross-validation setup, and smaller models degrade more under long runs due to limited context \cite{liu2024lost,zeng2024enhancing}. In the synthesis module, scale does not prevent hallucinations: reviewers can accept plausible but wrong abstracts, and even frontier models still hallucinate on technical QA, especially on multi-hop or conflicting evidence \cite{zheng2024judging,chern2024factool,min2023factscore}.\\
\textbf{The open weight verification tradeoff}: 
Closed-weight proprietary models (e.g., GPT-4, Claude 3.5 Sonnet) often outperform open-weight models but their closed design prevents inspection or modification of the reasoning process, which makes auditing, verification, and reproducibility difficult for autonomous research. In contrast, open-weight models can be fine-tuned to become competitive in low-data, low-resource, domain-specific settings, while remaining easier to adapt and control. They support stronger safety and governance hooks such as privacy-preserving training, mechanistic interpretability and better abstention behavior~\cite{wolfe2024laboratory}. \\
\textbf{Module-specific failure propagation}: Systematic evaluations of research agents report high failure rates across the full pipeline. In an analysis of 1,642 multi-agent traces, 44.2\% of failures were specification errors introduced during planning, and execution failures ranged from 41\% to 86.7\% depending on the architecture and whether constraints were persistently stored \cite{cemri2025mast}. Independent evaluation of \textit{The AI Scientist} further found that 42\% of proposed experiments failed to run \cite{arxiv:2502.14297}. Most notably, PaperBench reports that 100\% of agent-generated papers contained experimental or methodological weaknesses, and Claude 3.5 Sonnet achieved only 1.8\% task completion despite being a frontier model \cite{zhu2025aiscientists}.

The following sections formalize these requirements through the auditable autonomous research \textbf{AAR} standard (section~\ref{sec:audit}) quantifying when research agents achieve the verifiability necessary for autonomous scientific inquiry~\cite{souza2025provagent,fernsel2024auditability}, and \textbf{semantic provenance graphs} (section~\ref{sec:semantic-provenance}) providing the architectural substrate making auditable autonomous research tractable~\cite{moreau2013provenance,besta2025kgot}.

\section{The AAR standard: A measurement framework for auditable research}
\label{sec:audit}

In section~\ref{sec:failures}, we discussed three systematic failure modes in autonomous research agents: objective drift (experiments contradict their goals), transient constraints (specifications lost during execution), and unverifiable inference (citations decorrelated from claims). These failures share a common architectural deficit, i.e., the absence of persistent, queryable representations of reasoning that enable systematic verification. Current agents optimize for generation fluency while treating verification as an afterthought, producing systems that are impressively capable yet fundamentally untrustworthy for scientific work~\cite{lu2024aiscientist,arxiv:2502.14297}. This section establishes the \textbf{AAR standard} - a measurement framework defining what properties must be quantified to assess research agent trustworthiness. We argue that auditability must be measurable to be enforceable, and that current benchmarks' exclusive focus on task completion creates perverse incentives that reward unverifiable generation.
\subsection{The auditability invariant}
We begin with a foundational principle that any research-grade agent must satisfy:

\begin{definition}[Research-grade auditability]
A research agent is auditable if independent reviewers can verify claim correctness with effort $E_{\text{verify}} \ll E_{\text{generate}}$, where verification accesses only: (1) agent outputs, (2) structured provenance graphs, and (3) cited sources. Systems requiring re-execution of experiments or manual reconstruction of reasoning from logs fail this criterion.
\end{definition}

This definition immediately reveals why current systems fail. When independent evaluators identified the AI Scientist's energy efficiency paradox (section \ref{sec:inference}), they spent hours manually tracing through code and logs to reconstruct what the agent actually optimized for~\cite{arxiv:2502.14297}. This effort approached that of conducting the original research, defeating the purpose of automation.

\subsection{Four properties of auditable research}
Each documented failure mode implies a measurable property that, if monitored, would expose problems before outputs propagate to downstream use. We propose four complementary metrics:
\begin{definition}[Provenance coverage: Can claims be traced?]
The fraction -- $PCov$ -- of claims with complete traceable paths from sources through reasoning steps to outputs~\cite{souza2025provagent,moreau2013provenance}. High $PCov$ indicates comprehensive traceability; low $PCov$ indicates prevalence of orphaned claims lacking verifiable foundations.
\end{definition}
\begin{definition}[Provenance soundness: Do citations actually support claims?]
Semantic validity $PSnd$ measuring whether cited sources actually entail attributed claims, not merely whether citations resolve to real papers~\cite{chelli2024hallucination}.
$PCov$ is necessary but insufficient, i.e., citations may resolve to real papers yet not support the specific relationships asserted, creating scientific pollution: authoritative-looking text with real DOIs but severed logical links between evidence and claim.
\end{definition}
\begin{definition}[Contradiction transparency: Are evidence conflicts surfaced or suppressed?]
The proportion ($CTran$) of actual evidence conflicts that are detected and reported rather than suppressed through aggregation~\cite{wang2024factcheck,manakul2023selfcheckgpt}. Vector-based retrieval systems average contradictory embeddings, producing hedged summaries that obscure which specific sources disagree.
 \end{definition}
\begin{figure*}[!ht]
\centering
\includegraphics[width=\textwidth]{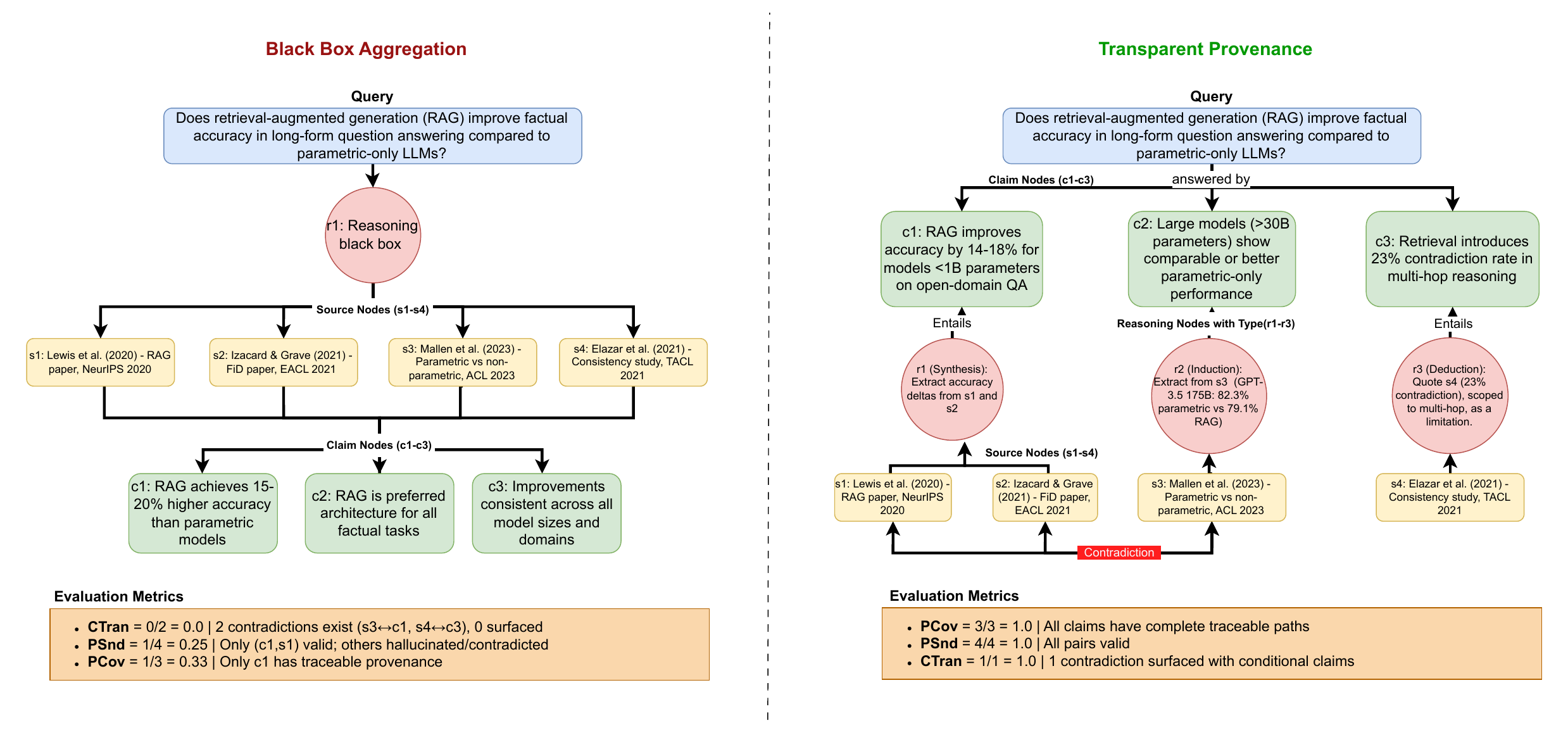}
\caption{Provenance tracking in RAG Systems. Black-box aggregation (left) versus transparent provenance (right) for identical sources. Black-box reasoning hides two contradictions ($s3 \leftrightarrow c1$, $s4 \leftrightarrow c3$), produces invalid citations, and leaves $c2/c3$ ungrounded -- yielding $CTran=0.0$, $PSnd=0.25$ (1/4 valid pairs with $\nu>0.5$), $PCov=0.33$. Explicit reasoning surfaces model-size contradiction, validates all citations and traces all claims -- achieving $CTran=1.0$, $PSnd=1.0$, $PCov=1.0$. Metrics computed using NLI entailment scores ($\nu$).}
\label{fig:graph_example}
\end{figure*}

\subsection{Formal metric definitions}

\textbf{Semantic provenance graph construction.} 
Given a user query $q$, a deep research agent constructs a directed provenance graph $G_q = (V, E)$ that traces how $q$ is answered through retrieved evidence and reasoning. For each query $q$, the agent executes a pipeline that retrieves a set of sources ($\mathcal{S}_q$), performs a sequence of reasoning steps ($\mathcal{R}_q$) that synthesize evidence from those sources, and generates a set of final claims ($\mathcal{C}_q$) that directly respond to the query $q$. The resulting provenance graph is then defined as ($G_q = (V, E)$), where the vertex set is the union of these components, i.e., $(V = \mathcal{S}_q \cup \mathcal{R}_q \cup \mathcal{C}_q)$.

\begin{definition}[Source nodes]
Each $s \in \mathcal{S}_q$ is a tuple ($id$, DOI/URL, excerpt, timestamp) representing external evidence retrieved for $q$. Source nodes have no incoming edges.
\end{definition}
\begin{definition}[Reasoning nodes]
Each $r \in \mathcal{R}_q$ is a tuple (id, inference, type, model) where \textit{type} $\in$ \{\text{deduction}, \text{induction}, \text{synthesis}\} and \textit{model} identifies the generating LLM.
\end{definition}

\begin{definition}[Claim nodes]
Each $c \in \mathcal{C}_q$ is a tuple \\($id$, statement, location) where \textit{statement} is the factual assertion answering (part of) $q$ and \textit{location} specifies its position in the output. Claim nodes have no outgoing edges.
\end{definition}

\begin{definition}[Typed edges]
Each edge $(u, v, \text{rel}, \nu) \in E$ connects nodes with relation \text{rel} $\in \{\text{supports}, \text{contradicts}, \text{refines}, \text{prerequisite}\}$. For \textit{supports} edges, $\nu \in [0,1]$ quantifies entailment strength; otherwise $\nu = \text{null}$.
\end{definition}

\begin{definition}[Provenance path]
For any claim $c \in \mathcal{C}_q$, the provenance path ($\pi(c)$) is defined as the subgraph obtained by traversing backward from $c$, where $V_c$ consists of all vertices $v \in V$ that have a directed path to $c$, and the source and reasoning subsets are given by $S_c = V_c \cap \mathcal{S}_q$ and $R_c = V_c \cap \mathcal{R}_q$, respectively.
This path traces how specific sources $S_c$ support claim $c$ through reasoning $R_c$, establishing the evidential lineage from $q$ to $c$.
\end{definition}

 We formalize the four \textbf{AAR} properties as computable functions over these sets.\\
\textbf{Provenance coverage}: The fraction of claims with complete provenance paths can be calculated as $PCov = \frac{|\{c \in \mathcal{C}: \pi(c) \text{ is complete}\}|}{|\mathcal{C}|}$\\
\textbf{Provenance soundness}: For each claim-source pair $(c, s)$ where $c$ cites $s$, define semantic validity $\nu(c, s) \in [0, 1]$ measuring whether $s$ entails $c$. Let $\mathcal{P} = \{(c, s) : c \in \mathcal{C}, s \in \mathcal{S}, c \text{ cites } s\}$ be all citation pairs.
\[
PSnd = \frac{1}{|\mathcal{P}|} \sum_{(c,s) \in \mathcal{P}} \mathds{1} [\nu(c, s) > \tau_{\text{entail}}]
\]
where $\tau_{\text{entail}}$ is a domain-specific threshold.\\ 
\noindent \textbf{Contradiction transparency}: Define $\mathcal{K} = \{(s_i, s_j, e) : s_i, s_j \in \mathcal{S}, e \in \text{entities}\}$ as the set of source pairs making contradictory claims about entity $e$. Let $\mathcal{D}$ be the subset of contradictions explicitly detected and reported by the system. Quantitatively, $CTran = \frac{|\mathcal{D}|}{|\mathcal{K}|}$.\\
\textbf{Audit effort}: This is measured empirically through human verification studies. Let $T_c$ be the time (minutes) required for a domain expert to verify claim $c$ given access to: (1) the claim, (2) provenance graph $\pi(c)$, (3) cited sources. The audit effort can be calculated as $AEff = \frac{1}{|\mathcal{C}_{\text{sample}}|} \sum_{c \in \mathcal{C}_{\text{sample}}} T_c$ where $\mathcal{C}_{\text{sample}}$ is a representative sample of claims. While $AEff$ measures human time, we can automate this score by simply counting the number of steps and sources in a claim's provenance path. Since the graph is already machine-readable, its structural complexity serves as a fast, automated proxy for how much effort a human would actually need.\\
\textbf{Measurement approach}: We assess auditability through the above four metrics. First, $PCov$ verifies that claims can be traced to their sources. Second, $PSnd$ uses formal verification to prove whether sources logically entail the attributed claims. Third, $CTran$ identifies contradictions by testing entailment bidirectionally -- checking whether evidence supports the claim as well as whether conflicting evidence exists. Fourth, $AEff$ measures verification time empirically through user studies with domain experts~\cite{mantravadi2025legalwiz,metropolitansky2025veritrail}. Persistently low \textbf{AAR} scores expose concrete architectural deficiencies -- low $PCov$ reflects missing provenance infrastructure, low $PSnd$ indicates weak semantic verification, low $CTran$ signals suppressed contradictions, and high $AEff$ denotes poor intelligibility -- making these limitations visible and actionable rather than obscured by aggregate performance metrics.\\
\noindent\textbf{Example.} For $q =$ ``Does retrieval-augmented generation (RAG) improve factual accuracy in long-form question answering compared to parametric-only LLMs?'', the agent retrieves $\mathcal{S}_q = \{s_1, s_2, s_3, s_4\}$ from the literature and generates output claims $\mathcal{C}_q = \{c_1, c_2, c_3\}$ where $c_1$ states ``RAG achieves 14-18\% higher accuracy than parametric models'', $c_2$ asserts ``RAG is preferred architecture for all factual tasks'', and $c_3$ claims ``Improvements generalize across model sizes and domains''. Figure~\ref{fig:graph_example} contrasts two architectures: \emph{black-box aggregation} (left) directly maps sources to claims without explicit reasoning, achieving $CTran=0.0$ (contradictions between $s_3 \rightarrow c_1$ and $s_4 \rightarrow c_3$ remain hidden), $PSnd=0.25$ (only 1 of 4 claim nodes has complete traceable paths), and $PCov=0.33$ (only $c_1$ is valid; $c_2$ and $c_3$ are hallucinated or contradicted). In contrast, \emph{transparent provenance} (right) exposes intermediate reasoning nodes $\mathcal{R}_q = \{r_1, r_2, r_3\}$ with typed inference (synthesis, induction, deduction), explicitly surfaces the contradiction, and achieves perfect auditability: $PCov=1.0$, $PSnd=1.0$, and $CTran=1.0$, enabling systematic verification of each claim's evidential support.

\section{Perspective: Semantic provenance with protocolized validation}
\label{sec:semantic-provenance}

We argue that auditable research agents need provenance graphs with protocolized validation. By this we mean systematic verification that measures entailment strength, source reliability, and extraction confidence according to explicit protocols, operating continuously during synthesis rather than as post-hoc review. Vector-based systems cannot reliably meet $PCov$, $PSnd$, $CTran$, and $AEff$ because embedding similarity does not represent entailment or contradiction. Similar wording can hide opposite claims, and aggregation can erase disagreements. We therefore require provenance that is persistent and queryable. It must record evidence support with measured strength (entailment strength, source reliability, extraction confidence). It must also represent conflicts and show how each conflict was resolved. Finally, it must store reasoning traces across decisions so reviewers can inspect why conclusions were chosen and test counterfactual changes. We show that post-hoc verification does not scale because it cannot recover missing reasoning chains and it arrives too late to prevent error build-up. Validation should run during synthesis with continuous feedback. We also integrate adversarial review into the provenance loop by writing validates/challenges edges that trigger retrieval, revision, or human escalation. Operational gating then blocks unverified claims from entering downstream outputs.

\section{Alternative views and objections}
\label{sec:objections}

The provenance-first architecture faces four recurring objections from practitioners prioritizing immediate deployment over verifiable research infrastructure. We address each systematically.

\noindent \textbf{Bigger models will solve this.} Scaling proponents argue that larger models will show emergent capabilities, making external provenance infrastructure unnecessary \cite{wei2022emergent,kaplan2020scaling}. We argue this claim conflates generation quality with attribution accuracy. Current systems still exhibit architectural amnesia across both open-weight and proprietary model families~\cite{arxiv:2502.14297}. More importantly, provenance is required even when the generated content is correct. If an agent synthesizes conclusions from dozens of papers, users must verify which specific sources support each claim, and this need is independent of model scale. Provenance also enables root-cause analysis when failures occur, supports iterative refinement through queryable audit trails, and provides reproducibility guarantees expected in scientific workflows \cite{souza2025provagent,herschel2017survey}.\\
\noindent \textbf{Graphs are too expensive.} A common objection is that provenance graphs are too expensive to maintain, especially for long-running research workflows, and that flat logs are a more practical option \cite{gehani2017practical}. However, recent evidence suggests that graph-based structure can be cost-effective in practice ~\cite{besta2025kgot,xu2025remindrag,bountris2025hyprov,bernal2024hipporag}. Knowledge graph of thoughts reports that graph-based task representations reduce cost while also improving success rates compared to stateless agents \cite{besta2025kgot}. \textsc{HippoRAG}~\cite{bernal2024hipporag} similarly shows that graph-based memory improves multi-hop reasoning that stateless retrieval often misses. Manual post-hoc validation can easily cost more than generation, while architectural validation enables automated graph checks that run in milliseconds per claim. At scale, research agents generate unreliable papers through hallucinated content, and this increases the risk of downstream reuse and training contamination~\cite{shumailov2024model,blau2024scientific}. For this reason, provenance graphs should be viewed as a cost control mechanism where they reduce verification effort, limit error propagation, and lower the long-term cost of maintaining a reliable scientific corpus.\\
\noindent \textbf{Logs provide sufficient traceability.} Some practitioners~\cite{bates2015linux} claim that timestamped logs provide enough traceability because they are easy to collect and have low overhead. We argue that logs only record actions and outputs; they do not capture why an agent makes a choice or how specific evidence supports a claim. For example, when The AI Scientist failed, the logs showed what ran, but they did not explain the key reasoning mistakes, which forced hours of manual inspection to piece together the decision path \cite{arxiv:2502.14297}. In contrast, semantic provenance graphs store explicit links from claims to evidence and reasoning steps, including how the system handles conflicting sources, which enables direct verification queries that flat logs cannot support~\cite{souza2025provagent,herschel2017survey}.\\
\noindent \textbf{Validation introduces prohibitive latency.}
Critics~\cite{koohestani2025agentguard,xia2024eddops,specmas2025neurips,amaral2024prove} argue that continuous verification adds latency and compute cost, and they claim this reduces the value of automation. They argue that when verification takes as long as generation, automation loses its efficiency advantage. We disagree because this view treats verification delay as the main cost and ignores larger downstream costs. A small verification delay saves much more time by preventing hallucinated reports that force long debugging and can contaminate later analyses. Independent evaluators of The AI Scientist spend expert effort similar to doing original research to find failures such as the energy efficiency paradox (section~\ref{sec:inference}).



\section{Conclusion}
\label{sec:conclusion}
Deep research agents are becoming capable of generating full scientific narratives at scale, so the limiting factor is no longer writing speed but whether outputs are verifiable before they pollute downstream science. This paper argues that today’s logs and citation practices are structurally insufficient because they record actions and references, not the claim-level evidence links needed to audit correctness. We therefore propose ``auditable-by-design'' research agents built around semantic provenance graphs that explicitly connect sources $\rightarrow$ intermediate reasoning $\rightarrow$ final claims in a persistent, queryable form, and we operationalize this goal with the \textbf{AAR} standard; provenance coverage (are claims traceable), provenance soundness (do sources actually support claims), contradiction transparency (are conflicts surfaced rather than averaged away), and audit effort (can a human verify faster than redoing the work). Framed this way, progress should be judged not only by task success but by whether verification is cheaper than generation, enabling scalable human oversight and safe autonomy. Looking ahead, the key engineering and research agenda is to standardize provenance schemas, improve entailment/contradiction checking over heterogeneous sources, and build benchmarks that reward low audit cost and high evidence fidelity so that agents produce not just plausible papers, but inspectable scientific artifacts.

\bibliographystyle{ACM-Reference-Format}
\bibliography{sample-base}


\end{document}